\documentclass{article}

\usepackage[preprint, nonatbib]{neurips_2023}

\usepackage[utf8]{inputenc} %
\usepackage[T1]{fontenc}    %
\usepackage{hyperref}       %
\hypersetup{
    colorlinks=true,
    linkcolor=red,
    citecolor=green,
    filecolor=magenta,      
    urlcolor=cyan,
}
\usepackage{url}            %
\usepackage{booktabs}       %
\usepackage{amsfonts}       %
\usepackage{nicefrac}       %
\usepackage{microtype}      %
\usepackage{xcolor}         %

\usepackage{amsmath}
\usepackage{nicefrac}       %
\usepackage{csquotes}
\usepackage{array}
\usepackage{adjustbox}
\usepackage{xspace}
\usepackage{multirow}
\usepackage{wrapfig}
\usepackage{import}
\usepackage{caption}
\usepackage{subcaption}
\usepackage{makecell}
\usepackage{enumitem}
\usepackage{gensymb}
\usepackage{wrapfig}
\usepackage{ragged2e}
\usepackage{algorithm,algpseudocode}
\usepackage{color, colortbl}
\usepackage{tabularx,booktabs}
\newcolumntype{Y}{>{\centering\arraybackslash}X}
\usepackage{bbm}
\usepackage{algorithm,algpseudocode}
\newcommand{\algrule}[1][.5pt]{\par\vskip.5\baselineskip\hrule height #1\par\vskip.5\baselineskip}

\def\ours{\texttt{\textbf{TABA}}\xspace}
\def\ie{\textit{{i.e.}}\xspace}

\makeatletter
\let\original@footnote\footnote
\newcommand{\align@footnote}[1]{%
  \ifmeasuring@
    \chardef\@tempfn=\value{footnote}%
    \footnotemark
    \setcounter{footnote}{\@tempfn}%
  \else
    \iffirstchoice@
      \original@footnote{#1}%
    \fi
  \fi}
\pretocmd{\start@align}{\let\footnote\align@footnote}{}{}
\makeatother

\newcommand\blfootnote[1]{%
  \begingroup
  \renewcommand\thefootnote{}\footnote{#1}%
  \addtocounter{footnote}{-1}%
  \endgroup
}

\usepackage{graphicx}

\graphicspath{{figures/}}

\title{Label Budget Allocation in Multi-Task Learning}

\author{Ximeng Sun$^{1\dagger}$ \ \ \ \  Kihyuk Sohn$^{2}$  \ \ \ \ Kate Saenko$^{1}$ \ \ \ \  Clayton Mellina$^{3\dagger}$ \ \ \ \  Xiao Bian$^{4\dagger}$\\
$^{1}$ Boston University, $^{2}$ Google Research, $^{3}$ Rarebase, $^{4}$ Roblox  \\
}

\begin{document}

\maketitle

\begin{abstract}
The cost of labeling data  often limits the performance of machine learning systems.~\blfootnote{$^\dagger$Work performed while at Google Cloud.} In multi-task learning, related tasks provide information to each other and improve overall performance, but the label cost can vary among tasks. \textit{How should the label budget (i.e. the amount of money spent on labeling)  be allocated among different tasks to achieve optimal multi-task performance?} We are the first to propose and formally define the label budget allocation problem in multi-task learning and to empirically show that different budget allocation strategies make a big difference to its performance. We  propose a Task-Adaptive Budget Allocation algorithm to robustly generate the optimal budget allocation adaptive to different multi-task learning settings. Specifically, we estimate and then maximize the extent of new information obtained from the allocated budget as a proxy for multi-task learning performance. Experiments on PASCAL VOC and Taskonomy  demonstrate the efficacy of our approach over other widely used heuristic labeling strategies. 
\end{abstract}

\section{Introduction}
In recent years, multi-task learning (MTL) focuses on simultaneously solving multiple related tasks and has attracted much attention~\cite{caruana1997multitask,misra2016cross,gao2019nddr,maninis2019attentive,sun2020adashare,yu2020gradient}.
By learning a shared representation across related tasks, MTL can significantly reduce 
training and inference time compared with single-task learning (STL),  while improving generalization performance and prediction accuracy~\cite{caruana1997multitask,Thrun98}.

In many real-world scenarios, data often plays a bigger role than training strategies in boosting the performance. However, data annotation is costly and sometimes restricted due to privacy concerns or a scarcity of experts. Given that practical applications are bound by a constrained label budget, it's essential to optimally allocate this budget among tasks to achieve the best multi-task performance. Recent MTL research
efforts mainly focus on reducing  negative transfer among tasks by designing novel network architectures~\cite{misra2016cross,gao2019nddr,ahn2019deep,sun2020adashare}, operating on gradients~\cite{chen2018gradnorm,yu2020gradient,maninis2019attentive} or clustering tasks into smaller groups~\cite{standley2020tasks,fifty2021efficiently}.  . Despite considerable advancements in MTL, the issue of how to allocate a limited label budget across tasks in the most efficient manner remains unexplored.

\begin{figure}[t]
\begin{center}
     \includegraphics[width=\linewidth]{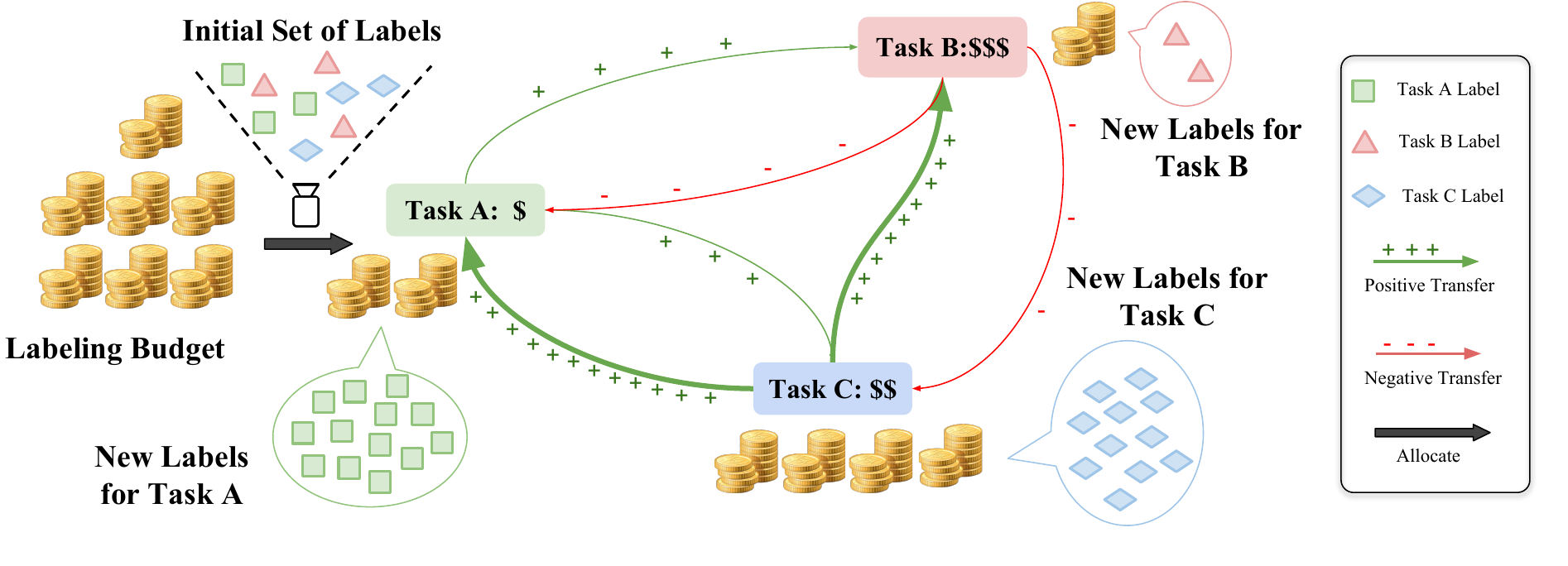}
\end{center} \vspace{-10pt}
    \caption{\small \textbf{Label Budget Allocation in MTL.}  A good budget allocation strategy should consider both the inter-task relationship and the different label costs. 
   }
   \label{fig:problem_definition} \vspace{-15pt}
\end{figure}

Motivated by this, we introduce and formalize the problem of label budget allocation in MTL. Suppose an initial set of labels is provided, along with a fixed amount of money is set aside for subsequent annotations. We pose the problem as follows: \textit{Without exceeding the annotation budget, how can we determine the number of additional labels to request for each task such that the multi-task learning system achieves the optimal overall performance?}
Logically, an effective budget allocation strategy in MTL should consider task relatedness and label costs of each task. In Figure~\ref{fig:problem_definition}, we exemplify this with a basic example  with Tasks A, B and C.
The bulk of budget should be spent on Task C as it provides the greatest benefits (i.e. positive transfer) to other tasks, while Task B should receive the least funding due to its detrimental effects (i.e. negative transfer) to other tasks and its high annotation cost.
In our preliminary study, we conduct an extensive grid search for the optimal budget allocation between two tasks on PASCAL VOC~\cite{Everingham15} and find that different budget allocations make a big difference in MTL performance. This demonstrates that the budget allocation problem deserves further research attention.

To address this problem, we propose the \textbf{T}ask-\textbf{A}daptive \textbf{B}udget \textbf{A}llocation (\ours) algorithm to generate robust and good budget allocation strategies which adapt to different task combinations, data distributions, label budgets, task-specific label costs and other configurations in MTL. \ours maximizes the total information gained from requested labels (used as a proxy of multi-task performance) with the given budget in two steps.
Firstly, \ours measures new information contained in each label type and estimates the extent of relative transferred information from one task's label to another task.
Secondly, \ours accounts for the declining rate of information gain from new labels when increasing labeled dataset size. With our estimated information gain, declining rate for each label type and the task-specific label costs set by users, \ours determines an optimal budget allocation via dynamic programming. In our experiment, we find that the performance of heuristic baselines is not robustness to the change of  cost ratios among tasks and MTL datasets. 
Experiments on PASCAL VOC~\cite{Everingham15} and Taskonomy~\cite{zamir2018taskonomy} have shown \ours consistently outperforms other widely used heuristic labeling strategies in various MTL settings.

The main contributions of our work are as follows:
\begin{itemize}[leftmargin=*]
    \item We are the first to define the label budget allocation problem in the context of multi-task learning 
    and empirically show that MTL performance is greatly impacted by different allocation strategies.
    \item We propose \ours algorithm to produce the optimal allocation adaptive to different MTL settings.
    \item We conduct experiments on PASCAL VOC and Taskonomy with different budgets and various task-specific label costs to demonstrate the superiority and robustness of our proposed approach over other widely used heuristic labeling strategies in MTL.
\end{itemize}

\section{Related Works}
\textbf{Multi-Task Learning.} Soft-parameter sharing MTL~\cite{misra2016cross,ruder2019latent,gao2019nddr} employs intermediate layers for task-specific backbones to facilitate task information exchange, while hard-parameter sharing MTL~\cite{huang2015cross,kokkinos2017ubernet,ranjan2017hyperface,bilen2016integrated,bragman2019stochastic,jou2016deep,dvornik2017blitznet} relies on shared hidden layers and task-specific branches. Some methods introduce task-specific attention branches~\cite{liu2019end,maninis2019attentive,ye2022inverted,vandenhende2020mti}, while others dynamically decide network sections to be shared across tasks~\cite{ahn2019deep,sun2020adashare,zhang2022automtl}. Beyond architectural considerations, numerous techniques seek to restrain negative transfer by lowering the statistical discrepancy between gradients from individual task losses. \cite{kendall2018multi,chen2018gradnorm} adaptively balance loss/gradient scales via task-specific loss adjustment, while others remove contradicting gradient components in tasks via gradient projection and regularization~\cite{yu2020gradient,suteu2019regularizing,navon2022multi,javaloy2021rotograd}. Adversarial training is used by some to render tasks' gradients similar~\cite{maninis2019attentive}. Complementing these studies, we investigate a new aspect in MTL: determining the optimal label budget allocation among tasks within a fixed MTL framework to enhance overall performance.

\noindent\textbf{Task Relatedness.}
Prior studies have explored feature sharing among tasks in shallow classification models through correlation of tasks' decision boundaries \cite{xue2007multi}, similarity of task-specific parameters \cite{jacob2008clustered,zhou2011clustered}, and evaluation of feature sharing patterns \cite{kumar2012learning,kang2011learning}. In deep learning, \cite{zamir2018taskonomy} and \cite{standley2020tasks} measure task relationships via transfer learning performance gains and task performance changes between STL and MTL, respectively. \cite{yu2020gradient,du2018adapting} suggest that cosine similarity of task gradients can indicate inter-task loss helpfulness. Task2Vec~\cite{achille2019task2vec} forms task vector representations from Fisher Information of model gradients. Nevertheless, these methods fall short in accurately gauging the relative informativeness of transferred labels, thereby lacking guidance for budget allocation. TAG~\cite{fifty2021efficiently} calculates task affinity by averaging task loss ratios before and after the gradient descent in the training step. Despite TAG's focus on training loss changes with varying task supervision, the loss function isn't always a reliable proxy for evaluation metrics~\cite{huang2019addressing,xu2018autoloss,wu2018learning,jenni2018deep}. Inspired by TAG, we propose a superior strategy for task relatedness computation for solving MTL label budget allocation, which directly based on evaluation metrics in a joint training context (see comparison with TAG in Sec.~\ref{sec:results}).

\noindent \textbf{Active Learning in Multi-Task, Multi-Domain Learning.} 
Active learning employs query rules, such as diversity~\cite{nguyen2004active,guo2010active,gal2017deep} or uncertainty-based rules~\cite{seung1992query,lewis1994sequential,tong2001support,joshi2009multi,beluch2018power,ranganathan2017deep}, to select unlabeled instances for labeling by an 'oracle'. \cite{reichart2008multi,zhang2010multi,saha2011online,fang2016active} integrate MTL with active learning, but are limited in task types and struggle to generalize across various deep learning vision tasks. In multi-domain learning, methods~\cite{li2012multi,zhang2016multi} actively label instances across domains without considering inter-domain relatedness. In this paper, we investigate label budget assignment to each task based on task relatedness, while randomly selecting unlabeled data for labeling. Active selection of unlabeled samples is earmarked for future research.

\noindent \textbf{Budget Allocation in Meta-Learning.} \cite{cioba2021how} addresses data budget allocation in meta-learning. It studies the uniform allocation of the fixed number of data points to various numbers of tasks in which each task receives the same amount of data. Without considering different costs of labels and the inherent task relationship,  \cite{cioba2021how} is generally different from the goal of this paper and their method is not applicable in solving label budget allocation problem in MTL.

\noindent\textbf{Low Data in STL.} 
Numerous STL techniques aim to improve performance in low-data situations. Weakly-supervised learning focuses on predictive models for fine-grained tasks with coarser supervision, such as image category labels for object detection~\cite{Wan_2018_CVPR,wan2019c,tang2018pcl} or video-sentence pairs for video moment retrieval~\cite{tan2021logan}. Semi-supervised learning~\cite{zhu2009introduction,berthelot2019mixmatch,sohn2020fixmatch} is another low-data solution, using both labeled and unlabeled data to enhance performance. Although low-data STL techniques (like weak or semi-supervision) could further boost performance, we concentrate on a distinct issue ``label budget allocation in MTL'' and avoid these methods for simplicity.

\section{Problem Formulation}
In this section, we first define the problem of label budget allocation in MTL (in Sec.~\ref{sec:original_formula}) and then demonstrate the importance of the defined problem using the PASCAL VOC dataset (in Sec.~\ref{sec:budget_allocation_matter}). 

\begin{wrapfigure}{r}{0.5\textwidth}
\vspace{-50pt}
  \begin{center}
    \includegraphics[width=0.48\textwidth]{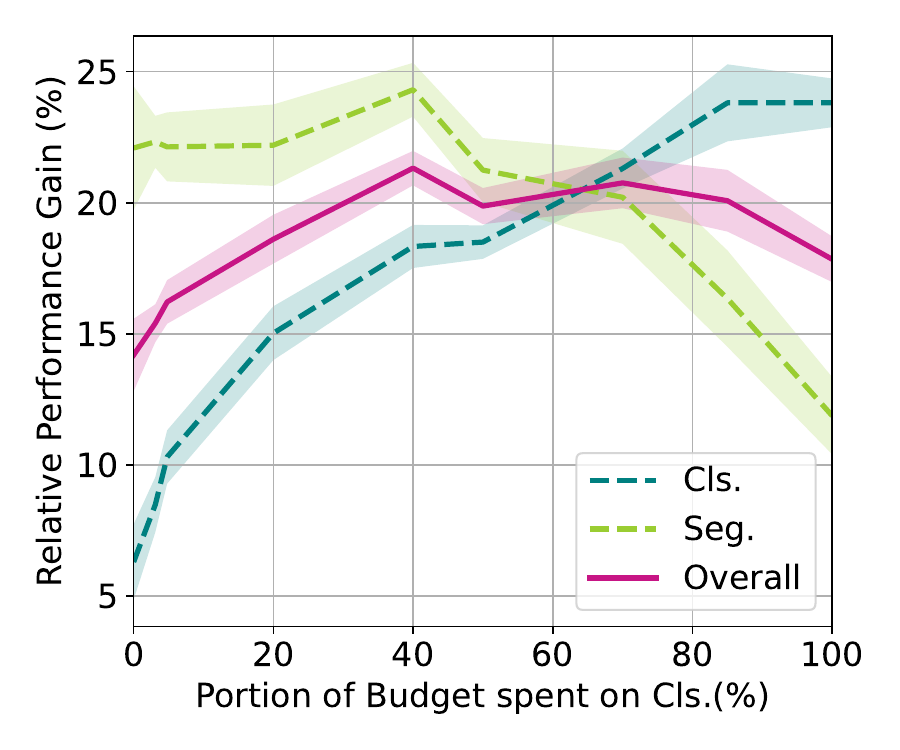}
  \end{center}
  \vspace{-10pt}
  \caption{\small On PASCAL VOC, we plot the relative single-task and multi-task performance with different allocations of the fixed label budget between two tasks. }  \label{fig:performance_vs_budget} 
  \vspace{-20pt}
\end{wrapfigure}

\subsection{Label Budget Allocation in MTL}\label{sec:original_formula}
Consider solving  $K$ tasks (the set of tasks is $T=\{\mathcal{T}_1, \mathcal{T}_2, \dots, \mathcal{T}_K\}$) jointly in multi-task learning.  For each task $\mathcal{T}_i$,  we provide $n$ labeled instances as the initial ``seed'' data. Suppose it costs $c_i$ to label a new instance in $\mathcal{T}_i$. Now, we are given $B$ units of budget to query additional ground-truth annotations from an `oracle' for the $K$ tasks. Within the total label budget $B$, our goal is to predict the number of new instances $N_i$ to label for each task $\mathcal{T}_i$ which will lead to the optimal MTL performance:
\begin{align}
    \text{argmax}_{N_1, N_2, \dots, N_K} & f (S_1, S_2, \dots, S_K), \label{eq:original_formula}\\
    & s.t. \sum_{i} N_i c_i \le B \nonumber,
\end{align}
where $S_i$ is the task $\mathcal{T}_i$'s performance and $f(\cdot)$ is a performance score aggregation function over multiple tasks. Without loss of generality, we treat all tasks equally (see the math form of $f(\cdot)$ in Sec.~\ref{sec:experimental_setup}). Note that we label the different instances in each task as it is a more general and  practical setting and it also outperforms labeling the same instances for all tasks (see Sec.~\ref{sec:results}).

\subsection{Does Label Budget Allocation Matter?}\label{sec:budget_allocation_matter}
We investigate the impact of budget allocation on individual task performance and overall MTL performance with PASCAL VOC~\cite{Everingham15} (In Fig.~\ref{fig:performance_vs_budget}) . We train the network to predict the present object categories, i.e. multi-label classification (\textit{Cls.}), as well as segment the image by objects, i.e. semantic segmentation (\textit{Seg.})~\cite{long2015fully,noh2015learning}. We pre-label 300 images/task as \enquote{seed} images, set $c_{cls.} = 1$ and  $c_{seg.} = 20$ based on the real label costs\footnote{\textit{e.g.}, \url{https://cloud.google.com/ai-platform/data-labeling/pricing\#labeling-costs}}, and allow 6300 units of label budget to request new labels\footnote{We set the budget as the total cost of labeling 300 new images for each task: $B = 300 \times (1 + 20)  = 6300$.}.

We vary the allocation of 6300 label budget between the two tasks. 
by gradually increasing the budget to request \textit{Cls.} labels (i.e. reducing the budget to request \textit{Seg.} labels). With the assigned budget, we randomly choose images to label in each task and re-train MTL network with both initial \enquote{seed} data and newly requested data. 
We evaluate each allocation via the performance gain of the new re-trained network over the original network only using initial seed data. We conduct the experiments for 10 runs independently with different newly requested data. From Fig.~\ref{fig:performance_vs_budget},
we observe:
\begin{itemize}[leftmargin=*]
    \item The MTL performance gain increases from $14.2\%$ to $21.3\%$ and then decreases to $17.8\%$ as the budget for multi-label classification increases from $0\%$ to $40\%$ to $100\%$ of the total budget. This demonstrates that  budget allocation among tasks has a significant impact on the overall MTL performance (maximum $7.1\%$ difference by only changing the budget allocation) and thus it is meaningful to look for an optimal budget allocation strategy in MTL.
    \item We achieve the best MTL performance gain ($21.3\%$) by spending $40\%$ of the budget on \textit{Cls.} and $60\%$ on \textit{Seg.}. It is different from heuristic labeling methods, such as spending the whole budget on one task ($14.2\%$ when spending all on \textit{Seg.} or $17.8\%$ all on \textit{Cls.}), labeling the same number of images for both tasks ($14.8\%$), or 
    equally splitting the budget between the two tasks ($19.8\%$).

     \item When we reduce the budget for \textit{Seg.} from $100\%$ to $0\%$, the \textit{Seg.} performance surprisingly increases at first and does not show a clear degradation until we reduce the \textit{Seg.} budget to $35\%$ of the total budget. To look into this positive transfer more closely, we take the equal allocation of budget on both tasks for example. 
     We further train a segmentation network with the same number of segmentation labels. The performance gain of segmentation drops rapidly from $21.4\%$ to $8.7\%$ after removing the \textit{Cls.}'s supervision. This  indicates that \textit{Cls.} has a significant positive transfer to \textit{Seg}. in this dataset, and that naively treating these tasks separately would not be ideal for budget allocation in MTL. We hence need to consider the information transfer among tasks when searching for an optimal allocation strategy.
\end{itemize}

\section{Our Approach: Task Adaptive Budget Allocation}

It is infeasible to perform a grid search for the optimal allocation of label budget and re-train MTL network with new requested labels every time as in Sec.~\ref{sec:budget_allocation_matter}. Therefore, 
there is a great need for an estimation of MTL performance from the number of new labeled images in each task without actually training the network. To this end, we propose the \textbf{T}ask \textbf{A}daptive \textbf{B}udget \textbf{A}llocation (\ours) algorithm to generate the optimal budget allocation strategy by maximizing a proxy of MTL performance. 

\begin{figure}[t]
\begin{center}
     \includegraphics[width=\linewidth]{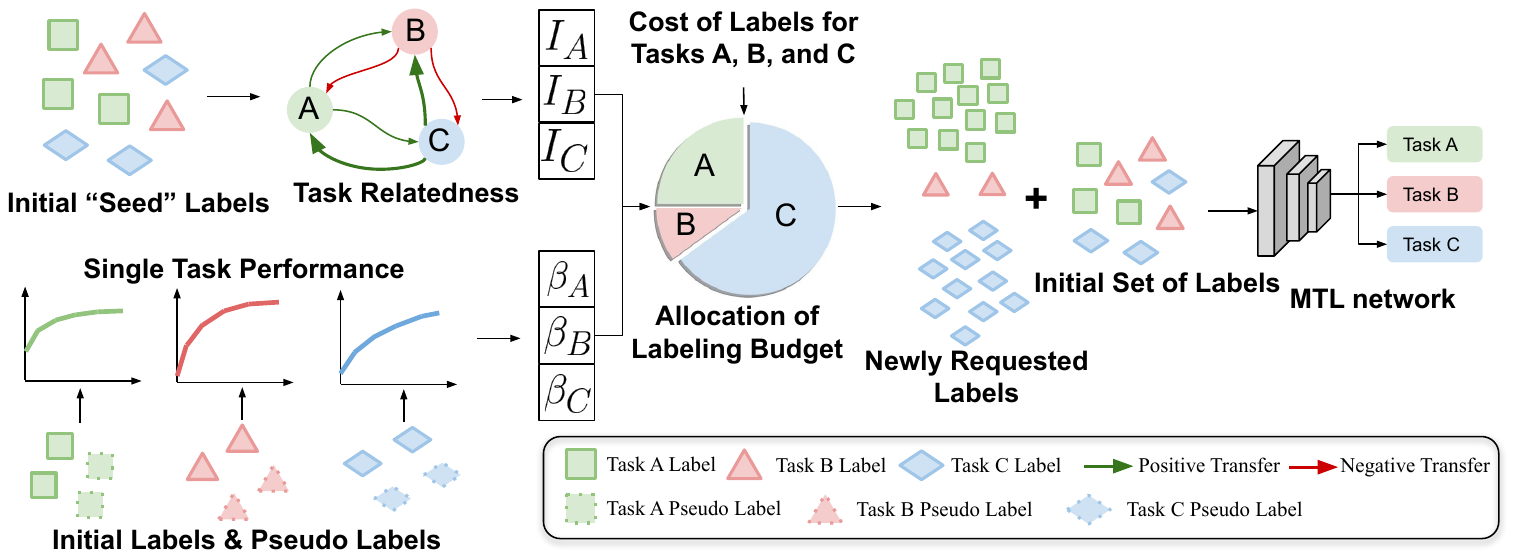}
\end{center} \vspace{-10pt}
   \caption{\small \textbf{Overview of TABA Algorithm.} We show the workflow of \ours via an example of  Tasks A, B and C. We measure the informativeness of each label type ($I_A$, $I_B$ and $I_C$) based on the task relatedness shown in the initial ``seed'' data. Meanwhile, we estimate the reduction rate of information gain for each task ($\beta_A$, $\beta_B$ and $\beta_C$) from STL performance on pseudo labels. Given $I$s, $\beta$s and the cost for each label type,  \ours produces a label budget allocation via dynamic programming. We re-train the MTL network with all newly requested labels and the initial labeled data.
   }
   \label{fig:flow_chart} \vspace{-15pt}
\end{figure}

\vspace{1mm}
\textbf{Approach Overview.}
Fig.~\ref{fig:flow_chart} illustrates an overview of \ours. We first measure the \textit{informativeness} $I_i$ provided by the label of task $\mathcal{T}_i$. It represents the contribution of one labeled instance to MTL performance. Then,  we gather the informativeness from newly requested labels as a proxy of the improvement of multi-task performance after spending the label budget. We compute the number of new labels $N_i$ for the task $\mathcal{T}_i$ by  maximizing the overall informativeness obtained from the budget:
\begin{align}
    \text{argmax}_{N_1, N_2, \dots, N_K} \sum_i{g_i(I_i, N_i)}, \label{eq:value_optimziation} \ \ \ \  s.t. \sum_{i} N_i c_i \le B 
\end{align}
where $g_i (\cdot)$ is the gathering function to compute the total informativeness of $N_i$ newly labeled images. 

In this paper, we measure the \textit{informativeness} $I_i$ as the integration of self-informativeness  $I_{i\rightarrow i}$, which is the label information to its own task,  and the transferred informativeness ($I_{i\rightarrow j}$, which is the transferred information of a label to all other tasks in the task group $T$:
\begin{align}
    I_i = I_{i \rightarrow i} + \sum_{j,  j\neq i} I_{i\rightarrow j}. \label{eq:v_i}
\end{align}
 Without loss of generality, we set self informativeness ($I_{i\rightarrow j}$), \ie a newly labeled image to its own task, as 1. As shown in Sec.~\ref{sec:budget_allocation_matter}, it is important to evaluate the extent of $I_{i\rightarrow j}$. We propose to compute the pair-wise task relatedness to estimate $I_{i\rightarrow j}$ (described in \textbf{Transferred Information among Tasks}).

With the increasing number of labels for one task, we further consider the deductive effect of a new label to its own task as well as to other tasks. Taking the same assumption of considering the overlap of information provided in $N_i$ labels \cite{cui2019classbalancedloss}, we define $g_i(\cdot)$ with the reduction rate $\beta_i$ as follows:
\begin{align}
    g_i(I_i, N_i) = \sum_{k = 1}^{N_i} (\beta_i) ^{k-1} I_i = \frac{1-\beta_i^{N_i}}{1-\beta_i} I_i,
\end{align}
where $\beta_i \in [0, 1]$ is the reduction rate of $I_i$ for task $\mathcal{T}_i$. Instead of using a constant $\beta$ for all tasks, we estimate the task-specific reduction rate $\beta_i$ via STL on pseudo labels (see \textbf{Estimate the Reduction Rate of Information}). 
We solve the constrained combinatorial optimization problem (\ie get the optimal $N_1$, $N_2$, $\cdots$, $N_k$ in Eq.~\ref{eq:value_optimziation}) with dynamic programming and randomly select $N_i$ unlabeled instances to request labels for each task. With the initial seed data and newly labeled data, we re-train the network jointly by aggregating gradients from all tasks.

\vspace{1mm}
\textbf{Transferred Information among Tasks. }
In a hard-parameter sharing MTL framework~\cite{huang2015cross,bragman2019stochastic,jou2016deep}, we update the shared backbone with the aggregated gradients from all tasks, where each task's information implicitly transfers, either positively or negatively, to all others.
We propose to estimate the relative transferred information in MTL based on the task relatedness revealed in the initial ``seed'' data. Specifically, when computing the transferred information from task $\mathcal{T}_i$ to task $\mathcal{T}_j$ at training step $t$ in MTL (i.e. $I^t_{i\rightarrow j}$), we probe the change of task $\mathcal{T}_j$'s performance (represented by its evaluation metrics) before and after jointly training with $\mathcal{T}_i$ for $m$ iterations:
\begin{align}
    I^t_{i\rightarrow j } = \frac{ S^{t+m}_j |_{(\mathcal{T}_i, \mathcal{T}_j)} - S^{t+m}_j |_{\mathcal{T}_j}} { S^{t+m}_j |_{(\mathcal{T}_j, \mathcal{T}_j)} - S^{t+m}_j |_{\mathcal{T}_j}},\label{eq:informative_score}
\end{align}
where $S_j^{t+m}|_{(\mathcal{T}_i, \mathcal{T}_j)}$,  $S^{t+m}_j |_{(\mathcal{T}_j, \mathcal{T}_j)}$ and $S_j^{t+m}|_{\mathcal{T}_j}$ are $\mathcal{T}_j$'s performance after joint training with tasks $\mathcal{T}_i$ and $\mathcal{T}_j$, trained with the aggregated gradients from the two mini-batches of task $\mathcal{T}_j$ and normally trained with only $\mathcal{T}_j$ respectively. Following \cite{fifty2021efficiently}, we evaluate $I^t_{i \rightarrow j}$ every 10 steps and take the average of $I^t_{i \rightarrow j}$ over the training as $ I_{i\rightarrow j}$.

\begin{figure}[t]
\begin{center}
     \includegraphics[width=\linewidth]{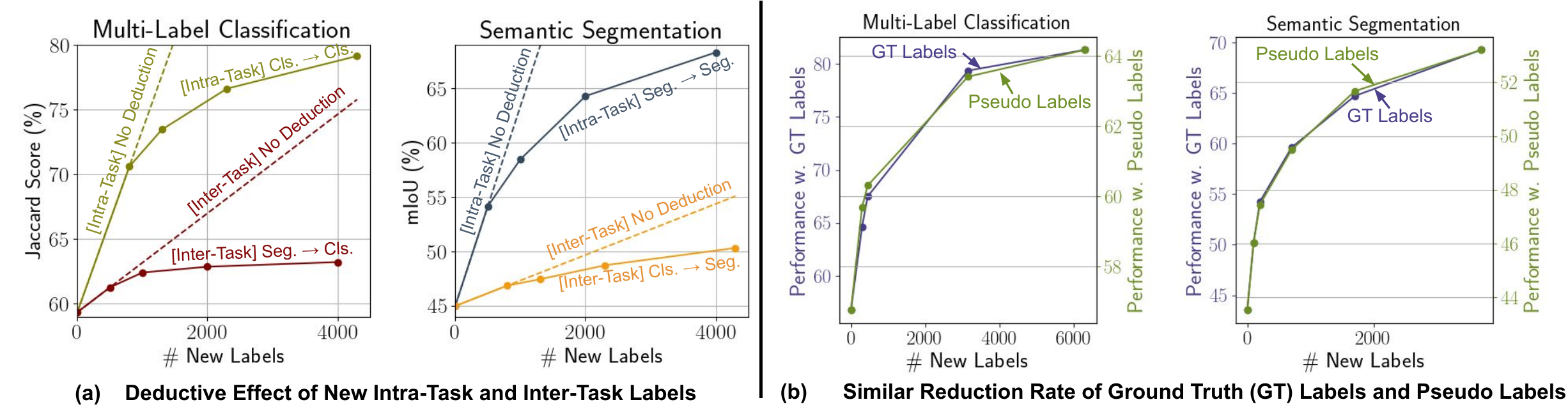}
\end{center} \vspace{-10pt}
   \caption{\small \textbf{Reduction Rate of Information.} In \textbf{(a)}, when we increase the amount of intra-task labels  or inter-task labels, the performance does not improve linearly. In \textbf{(b)}, we find that the performance curve of increasing ground truth labels matches the curve with increasing pseudo labels despite the two curves are in different scales.  
   }
   \label{fig:reduction_rate} \vspace{-15pt}
\end{figure}

\vspace{1mm}
\textbf{Estimate the Reduction Rate of Information. }
In the problem, we request $N_i$ (usually $N_i > 1$) new labeled images in each task $\mathcal{T}_i$.
With the increasing new labeled samples, new information of the same label type diminishes due to information overlap among data. In Fig.~\ref{fig:reduction_rate} (a), we clearly show the fading improvement of model performance when we keep increasing the amount of intra-task labels (\ie the labels for its own task) or inter-task labels (\ie the labels for other tasks). It indicates that the deductive effect of a new label both exists for its own task and other tasks in MTL. Following \cite{cui2019classbalancedloss}, we measure the overall informativeness of $N_i$ new labeled images by imagining adding them to the training set in a sequential manner–-one after the other.
Suppose that the information of each image does not overlap with another with probability $\beta_i$ for each task $\mathcal{T}_i$.  A new image does not provide any new information if its information is covered by any previous image. 
Let $\mathbbm{1}_k$ as a binary random variable that $\mathbbm{1}_k=1$ represents the $k^{\mathrm{th}}$ new image brings in new information and $\mathbbm{1}_k=0$ indicates the $k^{\mathrm{th}}$ image does not contain new information. We easily get $p(\mathbbm{1}_k = 1) = \beta_i^{k-1}$, where we call $\beta_i$ as the reduction rate of a new label for Task $\mathcal{T}_i$.
We define the expected overall information of $N_i$ new images in task $\mathcal{T}_i$ as:
\begin{align}
    g(I_i, N_i) & = \mathbb{E}[\sum_{k=1} ^{N_i} (\mathbbm{1}_k I_i)] = \sum_{k=1}^{N_i} p(\mathbbm{1}_k = 1) I_i = (\sum_{k=1}^{N_i} \beta_i^{k-1}) I_i = \frac{1-\beta_i^{N_i}}{1 - \beta_i} I_i. \label{eq:overall_value}
\end{align}
For simplicity, we do not consider partial or inter-task overlapping as well as difference between inter-task and intra-task reduction rates, which can be saved as an interesting future work. 

Furthermore, we find that the performance curve with the increasing ground truth labels matches the performance curve with the increasing pseudo labels despite the fact that they are in different scales (see Fig~\ref{fig:reduction_rate} (b)). It implies that the reduction rate of a new pseudo label is similar to that of a new ground-truth label. Motivated by this, we propose to estimate the task-specific $\beta_i$ by pseudo labels rather than treat $\beta_i$ as hyper-parameters in \cite{cui2019classbalancedloss}.  Concretely, we first predict pseudo-labels for the unlabeled images with an ensemble of STL networks trained with the initial ``seed'' set. We then compute the reduction rate of STL performance when expanding the training set with pseudo-labeled data. We define the initial performance gain from a pseudo-labeled image as $\Delta s_i$. Similar to Eq.~\ref{eq:overall_value}, the overall improvement $\Delta {S_i}$ of $N_i$ pseudo-labeled images is
\begin{align}
    \Delta S_i = \frac{1-\widehat{\beta_i}^{N_i}}{1 - \widehat{\beta_i}} \Delta s_i. \label{eq:overall_performance}
\end{align}
Without a closed form solution, we achieve the optimal $\widehat{\beta_i}$ and $\Delta s_i$ by minimizing the L$_1$ difference between $\Delta S_i$ and STL performance with gradient descent. From Fig~\ref{fig:reduction_rate} (b), we know that $\widehat{\beta_i}$ is a good approximation for $\beta_i$ while $I_i \neq \Delta s_i$. 

\vspace{1mm}
\textbf{Optimization of MTL Network. }
We jointly optimize the network with $K$ tasks. In an iteration, we sample different mini-batches from labeled instances of different tasks and aggregate the gradients from each task to update the shared parameters in MTL. The overall loss function is:
\begin{align}
    \mathcal{L}_{total} = \sum_{i=1}^{K} \lambda_i \mathcal{L}_i, 
\end{align}
where $\mathcal{L}_i$ represents the task-specific loss with task weighting $\lambda_i$. Note that it is required to set the same  $\lambda_i$ for  both computing task relatedness and the final MTL training with new acquired labels  since the task-relatedness depends on the loss weight $\lambda_i$.  In our experiments, we simply set $\lambda_i=1$ for all tasks while instead hyper-parameter sweeping can be done with the initial seed set before performing TABA algorithm.

\vspace{1mm}
\textbf{Efficiency of TABA.} 
The computation cost of TABA is durable compared to data annotation cost.  On PASCAL VOC dataset, we train the network which scores all tasks with 10 independent runs, which takes 25 GPU hours on Tesla A40 and we can compute the allocation of any amount of the label budget. According to \href{https://aws.amazon.com/sagemaker/data-labeling/pricing/}{the label cost} and \href{https://lambdalabs.com/service/gpu-cloud}{the cloud computing cost}, it takes less than \$50 to rent the machine for TABA algorithm while it takes \$252 to label the initial seed set and takes \$8889 to label the whole PASCAL VOC dataset. 
\section{Experiments}\label{sec:experiments}
We compare \ours with other widely used heuristic allocation strategies via experiments on PASCAL VOC~\cite{Everingham15} and Taskonomy~\cite{zamir2018taskonomy} with different multi-task learning settings. We provide the details of our experimental set-ups  and introduce our evaluation metrics for the labeling budget allocation problem. We show the superiority of \ours over baselines in Sec.~\ref{sec:results} and conduct ablation studies on the key components of \ours.

\subsection{Datasets and Tasks}\label{sec:datasets_and_tasks}

\textbf{PASCAL VOC 2-Task Setting.} PASCAL VOC is a collection of datasets for object detection. Following Deeplab-v3~\cite{chen2017rethinking}, we construct the training set by combining PASCAL VOC 2012~\cite{Everingham15} train split and additional annotations from SBD~\cite{BharathICCV2011} and test on PASCAL VOC 2012 validation set. 
The original dataset only provides semantic segmentation labels from which we extract class labels of the foreground objects in each image to form a multi-label classification task to predict the present object classes.
On PASCAL VOC, we optimize the network to learn two tasks, multi-label classification~($\mathcal{T}_1$) and semantic segmentation~($\mathcal{T}_2$), simultaneously.  

\noindent\textbf{Taskonomy 5-Task Setting.} The Taskonomy dataset~\cite{zamir2018taskonomy} includes over 4.5 million images from over 500 buildings with annotations for 26 different vision
tasks on every image ($\sim$12TB in size). Considering the limit of time and computing resources, we use the official 
tiny subset instead of the full dataset.  Following~\cite{standley2020tasks,sun2020adashare}, we sample 5 representative tasks out of 26 tasks for our experiments, namely semantic segmentation~($\mathcal{T}_1$), surface normal prediction~($\mathcal{T}_2$), depth prediction~($\mathcal{T}_3$), 2D keypoint estimation~($\mathcal{T}_4$) and 2D edge estimation~($\mathcal{T}_5$).

\subsection{Experimental Setup}\label{sec:experimental_setup}
\textbf{Label Budget Allocation Setup.} On PASCAL VOC, we randomly select 300 images to label ($n=300$) as the initial labeled set for each task. Since multi-label classification uses weak labels of semantic segmentation, we add the images with segmentation labels to multi-label classification training set in our experiments.  We carefully consult data-annotation and data-collection companies and set the per-image label cost 
$c_1:c_2 = 1:20$. To demonstrate the robustness of \ours with different costs, we vary the cost ratio between two tasks by setting $c_1:c_2=1:3$ (overpricing labels of multi-label classification) and $c_1:c_2=1:30$ (overpricing labels of semantic segmentation). In different scenarios, we keep the label budget as $B=300 \times (c_1 + c_2)$. 

On Taskonomy dataset, we randomly select 10,000 images ($n=10,000$) as the initial labeled set for each task. We set 
$c_1=10$, $c_2=3$, $c_3=3$, $c_4=2$,  $c_5=3$ \footnote{We refer to google and amazon labeling cost online and also consult data labeling companies (\ie Tolaka, Playment, Taqadam) for the labeling cost for each task. } and the label budget is $210,000$.

\noindent \textbf{Implementation Details.} We share the same backbone (DeepLab-ResNet-50)~\cite{chen2017deeplab} among all tasks. We use task-specific ASPP heads~\cite{chen2017deeplab} to generate predictions for all dense prediction tasks and use parallel FC layers for the multi-label classification task. Following \cite{yu2018slimmable,jin2020adabits,Sun_2021_ICCV}, we adopt the task-specific Batch Normalization (BN) Layers to resolve the discrepancy of data statistics in each task in the joint training. 
Note that \ours computes the informativemess based on evaluation metrics which do not raise specific requirement for the network architecture and can easily apply to other SOTA multi-task learning networks.
We use SGD optimizer with initial learning rate $0.001$ for the backbone and $0.01$ for ASPP heads and  cosine scheduler~\cite{loshchilov2016sgdr} to gradually anneal the learning rate to be 0.0001. We optimize PASCAL VOC for 100 epochs and Taskonomy for 30 epochs. 
Please refer to supplementary materials for more details. Our source code will be made publicly available.

\vspace{1mm}
\textbf{Baselines.}
In contrast to Sec.~\ref{sec:budget_allocation_matter}, it is unrealistic to perform the grid search over the whole allocation space to determine the best allocation strategy especially with the increasing number of tasks. On both datasets, we compare with three widely used heuristic practices to label instances for MTL: (1) 
we usually request the same amount of new images to label for each task (\enquote{Equal New Images}). We also compare requesting the labels for different images in different tasks and requesting the labels of the sames labels for all tasks (2) We spend all budget on a chosen task (\enquote{$100\%$ budget for one task}). 
(3) We distribute the budget equally to each task (\enquote{Equal Budget}). 

In our experiment, we first determine the number of new images to request labels by any given method (either a baseline or \ours).
Then, we randomly sample new images from the rest of training set, 
To reduce the randomness of data-sampling, we repeat sampling and re-training 10 times in PASCAL VOC and 5 times in Taskonomy and report the average performance.

\vspace{1mm}
\textbf{Evaluation Metrics.}
On PASCAL VOC, we evaluate multi-label classification via Jaccard Score\footnote{computes the intersection over union between the predicted object classes and the ground truth object classes.} and evaluate semantic segmentation with mIoU. In Taskonomy,  we follow~\cite{zamir2018taskonomy,standley2020tasks} and compute the task-specific loss on test images as the performance measurement for a given task. 

To effectively show the performance of the network trained with different budget allocations, we report the relative performance gain $\Delta_i$ for each task $\mathcal{T}_i$:
\begin{align}
    \Delta_i = (-1)^{l_i} (S_i - S_i') / S_i' \times  100\%,
\end{align}
where $S_i$ is the task performance score after adding new labeled data and $S_i'$ is the original performance score on the initial seed set. We set $l_i = 1$ if a lower value represents better performance for $S_i$ and 0 otherwise. Finally, we average over all tasks to get overall performance gain:
$ \Delta_{T}= f(S_1, S_2, \cdots, S_K) = \frac{1}{|T|}\sum_{i=1}^{K} \Delta_{i}.$
We report the relative performance gain in the main paper and direct
readers to refer to supplementary material for the performance of the original evaluation metrics (\textit{e.g.} mIoU, Jaccard Score).

\begin{figure}[t]
\begin{center}
     \includegraphics[width=0.9\linewidth]{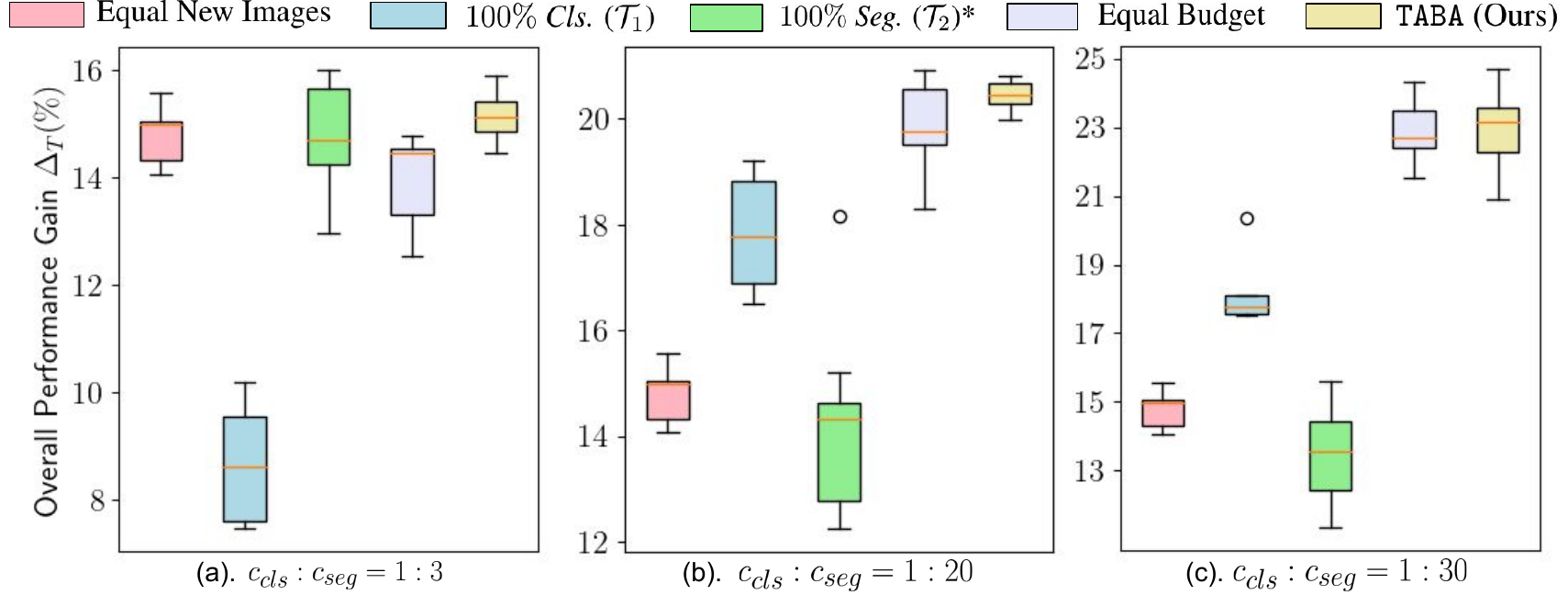}
\end{center} \vspace{-10pt}
   \caption{\small \textbf{Performance $\Delta_T$ on PASCAL VOC.} We evaluate our method and baselines with different cost ratios. In each scenario, the labeling budget is set as $B=300 \times (c_{cls} + c_{seg})$ so that \enquote{Equal New Images} is the same in (a), (b) and (c). * indicates providing labels of both tasks for the same group of images.
   }
   \label{fig:pascal_voc_main} \vspace{-10pt}
\end{figure}

\subsection{Results}\label{sec:results}

\begin{wraptable}{r}{0.5\linewidth} 
\vspace{-20pt}
  \begin{center}
  \caption{\small \textbf{Performance on PASCAL VOC with Larger Budget.} We increase the budget of  Fig.~\ref{fig:pascal_voc_main} (b)  to 9300, and keep $c_1:c_2=1:20$. * indicates providing labels of all tasks for the same group of images. }
        \resizebox{0.9\linewidth}{!}{
         \begin{tabularx}{1.1\linewidth}{ c| *{3}{Y}  }
            \Xhline{3\arrayrulewidth} 
            \multirow{2}{*}{Methods}  & \multicolumn{3}{c}{ $ c_1: c_2 = 1:20$ }  \\
            \cline{2-4}
             & $\Delta_{cls.}$ & $\Delta_{seg.}$ &  $\Delta_{T}$ \\
            \Xhline{3\arrayrulewidth} 
            Equal New Images & \cellcolor{red!5} $11.8_{(0.7)}$ & \cellcolor{red!5} $26.9_{(1.5)}$ & \cellcolor{red!5} $19.3_{(0.7)}$ \\
            $100\%$ \textit{Cls.} ($\mathcal{T}_1$)  & $26.6_{(0.3)}$ & $9.9_{(2.1)}$ & $18.3_{(1.2)}$ \\
            $100\%$ \textit{Seg.} ($\mathcal{T}_2$)* & $7.9_{(1.4)}$ & $24.5_{(2.8)}$ & $16.2_{(1.4)}$  \\
            Equal Budget &   \cellcolor{red!15} $21.8_{(0.5)}$ &  \cellcolor{red!15}  $23.9_{(2.0)}$ &  \cellcolor{red!15} $22.9_{(1.0)}$ \\
            \ours (Ours) &  \cellcolor{red!30}  $20.6_{(1.2)}$ & \cellcolor{red!30}  $28.9_{(2.2)}$ &  \cellcolor{red!30}  $24.7_{(0.6)}$ \\
            \Xhline{3\arrayrulewidth} 
        \end{tabularx}
        } 

\label{table:pascal_voc_larger_budget}
\end{center}
\vspace{-10pt}
\end{wraptable}

\textbf{PASCAL VOC.}  With \ours, we get $I_1=1.13$, $I_2=2.03$, $\beta_1=0.999$ and $\beta_2=0.997$ and we further compute the number of new images for each task by dynamic programming. 
Fig.~\ref{fig:pascal_voc_main} shows the performance of all baselines and our method with three different cost ratios between \textit{Cls.} and \textit{Seg.}. We keep the \enquote{Equal New Images} baseline unchanged by setting the label budget $B=300 \times (c_{cls} + c_{seg})$: \textbf{(1)} {By changing the cost ratios, the ranking of baselines are different.} For example, the best baseline is \enquote{Equal New Images} when $c_{cls}:c_{seg}=1:3$ while the best baseline is \enquote{Equal Budget} for the other two scenarios. Notably, \ours consistently outperforms all baselines with three different cost ratios which demonstrates the robustness of our method.  \textbf{(2)} 
{Devoting $100\%$ of the label budget to a single task fails to leverage this task relationship and results in a weaker improvement in the other task.} In contrast, \enquote{Equal Budget} and \enquote{Equal New Images} achieve more balanced improvement on both tasks. \ours not only improves both tasks in a well-balanced manner but also gets the best overall improvement. \textbf{(3)} {\ours achieves a larger improvement over all baselines with a higher budget (9300).}  
We further keep $c_{cls}:c_{seg} = 1:20$ (see Table.~\ref{table:pascal_voc_larger_budget}). \ours outperforms the second best method by $1.8\%$ in $\Delta_T$, demonstrating the importance of adaptively allocating the label budget even when the available budget grows.

\noindent\textbf{Taskonomy.} With \ours, we get $I_1=0.73, I_2=0.23, I_3=2.24, I_4=2.74, I_5=2.17$ and $\beta_1=0.9964, \beta_2=0.9997, \beta_3=0.9998,\beta_4=0.9999, \beta_5=0.9998$ for the given five tasks. Table~\ref{table:taskonomy_main} presents the results of \ours and all baselines in Taskonomy Dataset~\cite{zamir2018taskonomy}: \textbf{(1)} Unlike PASCAL VOC, the inter-task relationship is more complex within five tasks, and label budget allocation is thus more challenging in Taskonomy. When we spend the full budget on semantic segmentation ($\mathcal{T}_1$), the other four tasks perform worse, and conversely when we spend the full budget on each of the other four tasks, semantic segmentation performs worse. This reveals negative interference between the mid-level vision task ($\mathcal{T}_1$) and  the other low-level vision tasks ($\mathcal{T}_2 \sim \mathcal{T}_5$), which also correlates with the computed taskonomy graph in \cite{zamir2018taskonomy} in which semantic segmentation is far from other four tasks. \textbf{(2)} In contrast to the promising performance of \enquote{Equal Budget} baseline on PASCAL VOC, this baseline receives poor performance (ranked sixth among nine methods) on Taskonomy. Instead, allocating the full budget to keypoint estimation ($\mathcal{T}_4$) yields better performance than all other baselines. However, it causes unbalanced improvement over the five tasks and even impairs semantic segmentation performance by $8\%$ due to the negative transfer. (3) The allocation strategy generated by \ours achieves the best performance over all compared methods, yielding $4.8\%$ overall improvement over the initial \enquote{seed} set  performance with the given budget (v.s. $3.9\%$ by allocating all budget to keypoint detection). 

\begin{table*}

  \begin{center}
  \caption{\small \textbf{Performance on Taskonomy Dataset.}  \ours outperforms all widely used empirical baselines in Taskonomy with five task. * indicates providing labels of all tasks for the same group of images.}
        \resizebox{0.9\linewidth}{!}{
         \begin{tabularx}{ 1.1 \linewidth}{ c| *{6}{Y}  }
            \Xhline{3\arrayrulewidth} 
            \multirow{2}{*}{Methods}  & \multicolumn{6}{c}{ $B=210k, c_{seg}: c_{sn}: c_{depth}: c_{keypoint}: c_{edge} = 10:3:3:2:3$ }  \\
            \cline{2-7}
             & $\Delta_{seg}$ & $\Delta_{sn}$ & $\Delta_{depth}$ & $\Delta_{keypoint}$ & $\Delta_{edge}$ &  $\Delta_{T}$ \\
            \Xhline{3\arrayrulewidth} 
            Equal New Images &  \cellcolor{red!5}  $9.3_{(1.6)}$ &   \cellcolor{red!5} $4.7_{(0.2)}$ &  \cellcolor{red!5}  $4.0_{(0.8)}$ &  \cellcolor{red!5} $0.7_{(0.2)}$ & \cellcolor{red!5}  $0.6_{(0.1)}$ &  \cellcolor{red!5} $3.8_{(0.2)}$ \\
            Equal New Images* &  $5.5_{(0.03)}$ & $-7.4_{(0.1)}$ & $-5.9_{(1.5)}$ & $-4.3_{(0.2)}$ & $0.1_{0.3}$ & $-2.4_{0.3}$ \\
            $100\%$ \textit{Seg.} ($\mathcal{T}_{1}$)  & $12.2_{(2.4)}$ &  $-1.4_{(0.7)}$ &  $-0.4_{(1.6)}$ & $-0.6_{(0.3)}$ & $-0.8_{(0.9)}$ & $1.8_{(0.5)}$ \\
            $100\%$ \textit{SN}  ($\mathcal{T}_{2}$)  &  $-2.6_{(1.6)}$ &  $10.5_{(0.2)}$ & $5.6_{(0.7)}$ & $2_{(0.5)}$ & $1.5_{(0.1)}$ &  $3.4_{(0.5)}$  \\
            $100\%$ \textit{Depth} ($\mathcal{T}_{3}$)  & $-4.5_{(1.1)}$ & $7.3_{(0.2)}$ & $8.9_{(0.7)}$  & $2.9_{(0.1)}$ & $1.1_{(0.6)}$  & $3.1_{(0.3)}$ \\
            $100\%$ \textit{Keypoint} ($\mathcal{T}_{4}$) &  \cellcolor{red!15}  $-8.0_{(0.4)}$ & \cellcolor{red!15}   $9.9_{0.2}$ &  \cellcolor{red!15}  $10.3_{(1.1)}$ &  \cellcolor{red!15}   $4.8_{(0.4)}$ & \cellcolor{red!15}   $2.4_{(0.1)}$ & \cellcolor{red!15}  $3.9_{(0.5)}$  \\
            $100\%$ \textit{Edge} ($\mathcal{T}_{5}$) & $-2.0_{(0.7)}$ & $6.8_{(0.1)}$ & $4.3_{(1.9)}$ & $2.6_{(0.1)}$ & $0.7_{(0.1)}$ & $2.5_{(0.1)}$ \\
            Equal Budget & $3.7_{(0.6)}$ & $2.6_{(1.1)}$ & $1.8_{(1.3)}$ & $0.1_{(0.2)}$ & $0.1_{(0.8)}$ & $1.7_{(0.4)}$  \\
            \ours (Ours) & \cellcolor{red!30} $2.0_{(0.2)}$ &\cellcolor{red!30}  $6.9_{(0.6)}$ & \cellcolor{red!30}  $10.5_{(1.1)}$ & \cellcolor{red!30} $3.2_{(0.4)}$ & \cellcolor{red!30}  $1.3_{(0.1)}$ & \cellcolor{red!30}  $4.8_{(0.7)}$\\
            \Xhline{3\arrayrulewidth} 
        \end{tabularx}
        } 
\label{table:taskonomy_main}
\end{center}
\vspace{-20pt}
\end{table*}

\begin{table}
  \begin{center}
  \caption{\small \textbf{Ablation Studies on PASCAL VOC.} We ablate on two key components on PASCAL VOC dataset. In (a), we compare our model to the allocation strategy without considering cross-task relatedness and the allocation strategy using task relatedness computed by TAG method. In (b), we ablate our model with various manually chosen $\beta$s. In each scenario, the labeling budget is set as $B=300 \times (c_{cls} + c_{seg})$. }
        \resizebox{0.9\linewidth}{!}{
        \begin{tabularx}{ 1.1 \linewidth}{ c| *{3}{Y} |*{3}{Y} | *{3}{Y} }
            \Xhline{3\arrayrulewidth} 
            \multirow{2}{*}{Methods}  & \multicolumn{3}{c|}{ $c_{cls}: c_{seg} = 1: 3$ } & \multicolumn{3}{c|}{ $c_{cls}: c_{seg} = 1: 20 $} &  \multicolumn{3}{c}{ $c_{cls}: c_{seg} = 1: 30$ }  \\
            \cline{2-10}
             & $\Delta_{cls.}$ & $\Delta_{seg.}$ & $\Delta_{T}$ & $\Delta_{cls.}$ & $\Delta_{seg.}$ &  $\Delta_{T}$ & $\Delta_{cls.}$ & $\Delta_{seg.}$ &  $\Delta_{T}$ \\
            \Xhline{3\arrayrulewidth} 
             & \multicolumn{9}{c}{ (a). Ablation Studies on Task Relatedness} \\
            \Xhline{3\arrayrulewidth} 
            $I_1=I_2=1$ & 12.3 & 10.6 & 11.5 & 19.2 & 20.6 & 19.9  & 22.0 & 21.6 & 21.8 \\
            TAG~\cite{fifty2021efficiently} & 11.8  & 14.6 & 13.2 & 18.5 & 21.2 & 19.8 & 20.1 & 21.6 & 20.9 \\
            \ours (Ours) &  11.8 &  18.5 &  \textbf{15.1 }&  19.6 & 21.2 &  \textbf{20.4} & 20.4 &  25.5 &  \textbf{22.9 }\\
            \Xhline{3\arrayrulewidth} 
            & \multicolumn{9}{c}{ (b). Ablation Studies on Reduction Rate} \\
            \Xhline{3\arrayrulewidth} 
            $\beta_1 =\beta_2 = 1$ & 13.0 & 4.6 & 8.8 & 23.8 &  11.8  &  17.8 &   26.6 &   9.9 &  18.3  \\
            $\beta_1 =\beta_2 = 0.999$ &11.8 &  18.5 &  \textbf{15.1 } & 18.7& 21.0& 19.9 & 19.9 & 20.8 & 20.3 \\
            $\beta_1 =\beta_2 = 0.99$ & 9.3 & 19.9 & 14.6 &  10.6 & 20.6 & 15.6 & 10.7 & 21.0 & 15.8   \\
            \ours (Ours) &  11.8 &  18.5 &  \textbf{15.1 }&  19.6 & 21.2 &  \textbf{20.4} & 20.4 &  25.5 &  \textbf{22.9 }\\
            \Xhline{3\arrayrulewidth} 
        \end{tabularx}
        }
\label{table:pascal_voc_ablation}
\end{center}
\vspace{-20pt}
\end{table}

\textbf{Ablate on Task Relatedness.} 
To show the effectiveness of our measurement of the informative score of each label type, we first omit the transferred information among tasks and only consider the informative score to its own task ($I = 1$ for all tasks).  
From Table~\ref{table:pascal_voc_ablation}, \ours achieves better performance than the method with $I=1$, which shows it is important to consider cross-task informative scores when allocating label budget in MTL. Then we replace our task relatedness computation with loss-based TAG~\cite{fifty2021efficiently} method. The comparison TAG and \ours indiciates that computing the task relatedness based on the real evaluation metrics is more effective than loss-based task relatedness in the application of budget allocation.

\noindent \textbf{Ablate on Reduction Rate.}  Without considering the slowing growth of the new information when increasing the number of labeled data (\textit{i.e.} $\beta_1=\beta_2 = 1$ in Table.~\ref{table:pascal_voc_ablation}), Eq.~\ref{eq:value_optimziation} degrades to spending the full budget on the task with higher informative score per cost, which is multi-label classification in PASCAL VOC. It results in the unbalanced improvement of different tasks. Following \cite{cui2019classbalancedloss}, we manually select $\beta_i$ as 0.999 and 0.99 for all tasks. We find out that $\beta_i=0.999$ produces good results which indicates $0.999$ generally suits the reduction rate of information for most tasks~\cite{cui2019classbalancedloss} on PASCAL VOC. In \ours, we estimate the task-specific reduction rate via the single-task learning performance and further improves the overall performance $\Delta_T$ by $2.6\%$ when $c_{cls}:c_{seg}=1:20$ compared to using 0.999 as the common reduction rate.

\section{Conclusion}
In this paper, we are first to quantitatively define the general form of label budget allocation in multi-task learning and empirically show that MTL performance is greatly impacted by different allocation strategies under the same label budget.
We further present a novel framework for adaptively producing the optimal allocation in different multi-task learning setting. 
We propose the Task-Adaptive Budget Allocation (\ours) algorithm to produce the optimal allocation adaptive to different multi-task learning settings. We demonstrate the efficacy of our proposed method via extensive experiments on PASCAL VOC and Taskonomy with different label budgets and various task-specific label costs. The allocation strategies generated by \ours consistently outperform other widely used heuristic budget allocation strategies. Moving forward, we would like to explore querying rules to select instances in each task to request labels and revisit active multi-task learning in the context of deep learning. Moreover, we are hoping to attract more research attention to this practical problem in the community and expect better solutions in the future.

\vspace{1mm}
\textbf{Limitations.} We experiment our method \ours based on the hard-parameter sharing MTL. Even though \ours does not raise any requirement for the model architecture, it might need specific tuning to adapt to other MTL frameworks. Also, random labeling after determining the number of labels in \ours can probably restrict the MTL performance and it would be interesting to replace it with  suitable active learning methods in the future.

\vspace{1mm}
\textbf{Broader Impacts.} Negative impacts of our research are difficult to predict, however, they are likely to be the usual pitfalls associated with deep learning models. These include susceptibility to adversarial attacks and data poisoning, dataset bias, and lack of interpretability. Other risks associated with the deployment of computer vision systems include privacy violations when images are captured without consent, or used to track individuals for profit, or increased automation resulting in job losses. While we believe that these issues should be mitigated, they are beyond the scope of this paper. Furthermore, we should be cautious of potential failures of the approach which could impact the performance/user experience of any high-level AI systems based on our research.

\clearpage
\appendix
\section{Algorithm}
We provide the detailed steps of \ours in Algorithm~\ref{alg:our_method} in addition to Fig. 3 in the main paper.  We measure the informativeness of each label type ($I_i$) with the initial ``seed'' data. Meanwhile, we estimate the reduction rate of information gain for each task $\beta_i$ from STL performance on pseudo labels. Given $I$s, $\beta$s and , label costs and the total label budget,  \ours produces a label budget allocation via dynamic programming. We train the MTL network with a new training set containing all newly requested labels and the initial labeled data.

\begin{algorithm}[H]
 \caption{\ours: Task-Adaptive Budget Allocation}~\label{alg:our_method}
\begin{algorithmic}[1]
    \Require
      \Statex $n$ labeled \enquote{seed} instances for $K$ tasks respectively
      \Statex Label Cost $c_i$ for Task $\mathcal{T}_i$ and Total Label Budget $B$
    \algrule
    
    \Statex \textbf{Part 1: } Compute the Informativeness $I_i$
    \For{$t = 1, \cdots, \text{iterations}$}
        \State Train MTL network with $K$ tasks
        \If{$t$ mod 10 = 0}
            \State $\forall i, j$,  compute $I^t_{i \rightarrow j}$
        \EndIf
     \EndFor
    \State Get  $I_{i \rightarrow j}$ by averaging $I^t_{i \rightarrow j}$ and get $I_i$ with Eq.(4)
    \algrule
     \Statex \textbf{Part 2: } Estimate the Reduction Rate $\beta_i$
        \State Train multiple STL networks with $n$ labeled \enquote{seed} instances
        \State Generate pseudo labels for unlabeled instances via the ensemble of STL networks
        \For{$a = a_1, a_2, a_3, \cdots, a_p$}
            \State Train STL network with $n$ labeled instances and $a$ pseudo-labeled instances
            \State Get the evaluate metrics $S_{i, a+n}$ on test set for each task $\mathcal{T}_i$
        \EndFor
        \State Estimate $\beta_i$ with $S_{i, a+n}$ 
    \algrule
    \State Compute $N_i$ via Dynamic Programming with $c_i$, $I_i$, $\beta_i$ and $B$.
    \State Randomly sample $N_i$ instances to get labels for Task $\mathcal{T}_i$
    \State Train MTL network with $n$ \enquote{seed} data and new labeled data
  \end{algorithmic} 
\end{algorithm}

\section{Experimental Setups}
There are two kinds of annotations widely used in computer vision: (1). Labels can be manually annotated, such as semantic segmentation, object classes. (2). Labels are generated by the camera itself, such as depth information. In our paper, we set  the cost of manual labels according to its data annotation cost. We set the cost of camera-generated labels as its data collection cost, since we need to collect images with specific camera requirements for camera-generated labels instead of web search. In the implementation, we use dilated convolution layers in DeepLab-ResNet-50 backbone with strides $[1, 2, 1, 1]$ and dilations $[1, 1, 2, 4]$ for four residual blocks respectively. We use ASPP head as the task specific heads for dense-prediction tasks, such as all tasks in Taskonomy and use a convolution layer followed by a fully connected layer as the task-specific head for multi-label classification. We optimize PASCAL VOC for 100 epochs. We optimize Taskonomy for 30 epochs on the full tiny set and extend the learning for the small training set until observing the converged validation loss. 
\section{Comparison in Real Evaluation Metrics}
In this section, we provide the comparison of all real evaluation metrics in PASCAL VOC~\cite{Everingham15} and Taskonomy~\cite{zamir2018taskonomy}(see Table \ref{table:pascal_voc_main}.-\ref{table:taskonomy_main}).

\begin{table*} [t]

  \begin{center}
        \resizebox{\linewidth}{!}{
        \begin{tabularx}{ 1.24 \linewidth}{ c| *{2}{Y} |*{2}{Y} | *{2}{Y} }
            \Xhline{3\arrayrulewidth} 
            \multirow{2}{*}{Methods}  & \multicolumn{2}{c|}{ (a) $c_{cls}: c_{seg} = 1: 3$ } & \multicolumn{2}{c|}{ (b) $c_{cls}: c_{seg}  = 1: 20 $} & \multicolumn{2}{c}{ (c) $c_{cls}: c_{seg}  = 1: 30$ }  \\
            \cline{2-7}
             & ${cls. \uparrow}$ & ${seg. \uparrow}$ & ${cls. \uparrow}$ & ${seg. \uparrow}$ & ${cls. \uparrow}$ & ${seg. \uparrow}$ \\
            \Xhline{3\arrayrulewidth} 
            Initial \enquote{Seed} Set & 65.92 & 46.33 & 65.92 & 46.33 & 65.92 & 46.33 \\
            Equal New Images & 72.17 & 55.65 & 72.17 & 55.65 & 72.17 & 55.65\\
            $100\%$\textit{ Cls.} ($\mathcal{T}_{1}$) & 74.47 & 48.44 & 81.61 & 51.83 & 83.47 & 50.91\\
            $100\%$ \textit{Seg.} ($\mathcal{T}_{2}$)& 70.79 & 56.58 & 70.07 & 56.55 & 70.46  & 55.61 \\
            Equal Budget & 72.45 & 54.53 & 78.12 & 56.14 & 80.34 & 54.81 \\
            \ours (Ours) &  73.71 & 54.88 & 78.88 & 56.15 & 79.36 & 58.14 \\
            \Xhline{3\arrayrulewidth} 
        \end{tabularx}
        } 
\caption{\small \textbf{Performance on PASCAL VOC.} We evaluate our method and baselines with different cost ratios between two tasks. Our method consistently outperform all baselines in all scenarios. In each scenario, the labeling budget is set as $B=300 \times (c_{cls} + c_{seg})$ so that \enquote{Equal New Images} is the same in (a), (b) and (c).}
~\label{table:pascal_voc_main}
\end{center}
\end{table*}

\section{Visualize Task Relatedness}
In Table~\ref{table:taskonomy_task_relatedness}, we visualize the task relatedness in the form of $I_{i \rightarrow j}$ averaged over five independent runs on Taskonomy dataset. In this paper, we only consider the pair-wise task transfers without counting the higher order transfer.

\begin{table*}
  \begin{center}
        \resizebox{0.9\linewidth}{!}{
        \begin{tabularx}{ 1.1 \linewidth}{c| c| *{5}{Y}  }
            \Xhline{3\arrayrulewidth} 
            \multicolumn{2}{c|}{} & \multicolumn{5}{c}{Task $\mathcal{T}_j$} \\
            \cline{3-7}
             \multicolumn{2}{c|}{} & ${seg }$ & ${sn }$ & ${depth }$ & ${keypoint }$ & ${edge }$   \\
            \hline
            \multirow{5}{*}{Task $\mathcal{T_i}$}  & ${seg }$ &  \cellcolor{green!100}  1.00 & \cellcolor{green!32}   0.32 & \cellcolor{red!37}   -0.37 & \cellcolor{red!21}   -0.21 & 0.00  \\ 
            &${sn }$ & \cellcolor{red!51}  -0.51 & \cellcolor{green!100} 1.00 & \cellcolor{green!16} 0.16 & \cellcolor{red!20} -0.20 & \cellcolor{red!20} -0.20  \\
            & ${depth }$ & \cellcolor{green!92}  0.92 &  \cellcolor{green!72} 0.72 &  \cellcolor{green!100} 1.00 & \cellcolor{red!30} -0.30 & \cellcolor{red!9} -0.09 \\
            &  ${keypoint }$ & \cellcolor{green!92} 0.92 & \cellcolor{green!51}  0.51 & \cellcolor{green!20} 0.20 & \cellcolor{green!100} 1.00 & \cellcolor{green!11} 0.11 \\
             & ${edge }$ &\cellcolor{green!57} 0.57 & \cellcolor{green!100} 1.12 & \cellcolor{red!33}  -0.33 & \cellcolor{red!19}  -0.19 &\cellcolor{green!100}  1.00 \\
            \Xhline{3\arrayrulewidth} 
        \end{tabularx}
        } 
        
\caption{\small \textbf{Taskonomy Task Relatedness.}  We mark the positive transfer in green and negative transfer in red.}
\label{table:taskonomy_task_relatedness}
\end{center}
\end{table*}

\end{document}